# Even vertex $\zeta$- graceful labeling on Rough Graph

R.Nithya[1], K.Anitha[2,*]

[1] Research Scholar, Department of Mathematics, SRM Institute of Technology, Ramapuram, Chennai-600089.

[2] Department of Mathematics, SRM Institute of Technology, Ramapuram, Chennai-6000089.

*Corresponding Author: anithak1@srmist.edu.in

## Abstract :

Rough graph is the graphical structure of information system with imprecise knowledge. Tong He designed the properties of rough graph in 2006[6] and following that He and Shi introduced the notion of edge rough graph[7]. He et al developed the concept of weighted rough graph with weighted attributes[6]. In this paper, we introduce a new type of labeling called Even vertex $\zeta$- graceful labeling as weight value for edges. We investigate this labeling for some special graphs like rough path graph, rough cycle graph, rough comb graph, rough ladder graph and rough star graph.

## 1.Introduction:

Rough set theory is a new mathematical tool for solving uncertain problems through indiscernibility between objects. It was proposed by Pawlak in 1982. Rough set theory's core premise is based on lower and upper approximations[3]. In rough set theory, a major important concept is rough membership function[4]. It has a wide range of application in the field of knowledge discovery and data mining, business, image processing, conflict analysis, decision making process etc. He T and Shi K introduced rough graph using binary relations and its structure[5]. T He, Y Chen, and K Shi first established weighted rough graph in 2006 using the class weights for the edge equivalence class[6]. In 2011, M.Liang, B.Liang, L.Wei and X.Xu defined edge rough graph based on the edge set partitioning of universe. Many researchers have described the properties of rough graph and their representation forms[7]. Following them Chen, Jinkun and Jinjin Li described an application of rough sets to graph theory[8]. Chellathurai and Jesmalar defined a weighted rough graph and its properties[9]. Bibin Mathew et al recently defined a vertex rough graph and its mathematical properties[20] and Metric dimension of rough graphs was introduced by Anitha and Arunadevi[19].

Labeling of a graph G is an assignment of integers to the vertices of G or edges of G or both satisfying certain conditions. A survey of graph labeling is produced by Gallian[2]. In 1967, Rosa introduced a labeling of G called β-valuation, later on Solomon W. Golomb called as graceful labeling which is an injection f from the set of vertices V(G) to the set {0, 1, 2, . . . , q} such that when each edge e = uv is assigned the label | f (u) − f (v)|, the resulting edge labels are distinct. A graph which admits a graceful labeling is called a graceful graph[2]. Grahamad Sloane proved that most graphs are not graceful. In 1991,Gnanajothi introduced a labeling of G called odd graceful labelling[2]. In 1985, Lo introduced a labeling of G called edge graceful labelling[2]. In 2009, Solairaju and Chithra introduced a labeling of G called edge odd graceful labelling [2]. Likewise, S.N.Daoud proved some of the paths and cycles are edge odd graceful labeling in 2017[11]. Solairaju et al defined various labelings for ladder[12] and Mohamed R.Zeen El Deen introduced Edge $\delta$- graceful labeling for some cyclic related graphs[13] and also he proved some results in edge even graceful labeling of the join of two graphs[14].

Md Forhad Hossain et al discussed new classes of graceful trees[18] and Md Shahbaz Aasai and Md Asif et al described radio labelings of lexicographic product of some graphs[17]. Hennig Fernaua et al introduced sum labeling for the generalised friendship graph[15] and Abdullah Zhraaa and Arif Nabeel et al introduced dividing graceful labeling for certain trees[16].In this paper, we introduced new type of labelling based on graceful and we investigate it in some rough graphs.

## 2.Preliminaries:

### 2.1 Rough Membership function[10]:

Assume $\mathbb{M} = (U, F)$ is an information system, $\emptyset = G \subseteq U$. In rough terms, here is the membership function for the set $\omega_G^F = \frac{|[f]_G \cap G|}{|[f]_F|}$ for some $f \in U$.

### 2.2 Rough graph[10]:

Let $\mathbb{U} = \{V, E, \omega\}$ be a triple consisting of non empty set $V = \{v_1, v_2, \dots v_n\} = \mathbb{U}$, where $\mathbb{U}$ is a universe, $E = \{e_1, e_2, \dots e_n\}$ be a set of unordered pairs of distinct elements of $V$ and $\omega$ be a function $\omega: V \to [0,1]$. A Rough graph is defined as

$$(v_i, v_j) = \begin{cases} max\left(\omega_G^V(v_i), \omega_G^V(v_j)\right) > 0, \text{edge} \\ max\left(\omega_G^V(v_i), \omega_G^V(v_j)\right) = 0, \text{no edge}. \end{cases}$$

**Example:**

Let U be an information system. Let U={1,2,3,4,5,6,7} and the set X={ 1,4,5,7}. Let age, hypertension and complication are condition attributes and delivery be the decision attribute. The information table is given as follows:

|      | Attributes |              |              | Decision |
| ---- | ---------- | ------------ | ------------ | -------- |
| Case | Age        | Hypertension | Complication | Delivery |
| 1    | 20…..29    | No           | None         | Fullterm |
| 2    | 20…..29    | Yes          | Obesity      | Preterm  |
| 3    | 20…..29    | Yes          | None         | Preterm  |
| 4    | 20…..29    | No           | None         | Fullterm |
| 5    | 30…..39    | Yes          | None         | Fullterm |
| 6    | 30…..39    | Yes          | Alcoholic    | Preterm  |
| 7    | 40…..50    | No           | None         | Fullterm |

R{1}={1,4}; R{2}={2}; R{3}={3}; R{4}={1,4}; R{5}={5}; R{6}={6}; R{7}={7}

Rough Membership values are

$$\omega_G(1) = \frac{|\{1,4\} \cap \{1,4,5,7\}|}{|\{1,4\}|} = \omega_G(4)$$

$\omega_G(2) = 0, \quad \omega_G(3) = 0, \quad \omega_G(5) = 1, \quad \omega_G(6) = 0, \quad \omega_G(7) = 1,$

The Rough graph is constructed for the above information table:

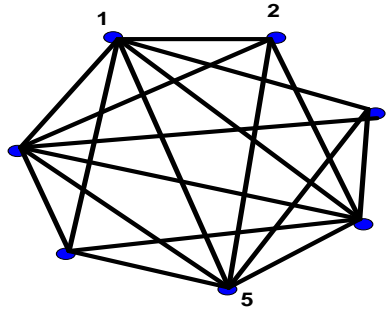

### 2.3 Rough Path graph[10]:

Distinct edges in Rough walk is said to be rough trail and distinct vertices in a rough walk is said to be rough path. It is denoted by $\mathcal{P}_n$.

### 2.4 Rough Cycle graph[10]:

A Rough cycle is defined as the closed Rough walk $v_1, v_2, \ldots v_n = v$ where $n \geq 3$ and $v_1, v_2, \ldots v_{(n-1)}$ are distinct. It is denoted by $\mathcal{C}_n$.

### 2.5 Rough Ladder graph[10]:

The Ladder Rough graph is defined as the Rough Cartesian product of Rough path and the Complete Rough graph. It is denoted by $\mathcal{L}_n$.

## 3. *Main Results:*

### 3.1 Rough Labeling Graph:

A Rough graph $R^\varphi(G) = (V^\varphi, E^\varphi, \omega)$ is said to be rough labeling graph if $V^\varphi = \{v_1^\varphi, v_2^\varphi, \ldots v_n^\varphi\}$ and $\omega: V^\varphi * V^\varphi \to [0,1]$ is injection such that the labeling of edges and vertices are distinct and there exists edge if $R^\varphi(G) = \max(\omega(v_i^\varphi), \omega(v_j^\varphi)) > 0$ for $v_i, v_j \in V$.

### *3.2 Even vertex $\zeta$ -graceful labeling:*

A function is called Even vertex $\zeta$- graceful labeling of a graph $G(V,E)$ with $n$ vertices and $m$ edges if $f: V(G) \to \{2,4,6 \ldots\}$ is bijection and the induced function $f: E(G) \to N$ are distinct and it is defined as $f^*(uv) = \begin{cases} \frac{\zeta}{2} & \zeta \text{ is even} \\ \frac{\zeta+1}{2} & \zeta \text{ is odd} \end{cases}$ where $\zeta = f(u) + f(v) + m(G)$ for all $u,v \in E$ are all distinct.

### *3.3 Rough $\zeta$ – graceful graph:*

A Rough graph $R^\varphi(G) = (V^\varphi, E^\varphi, \omega)$ has $n$ vertices and $m$ edges if $V^\varphi = \{v_1^\varphi, v_2^\varphi, \ldots v_n^\varphi\}$ and $\omega^\varphi: V^\varphi * V^\varphi \to [0,1]$ is bijection such that the labeling of vertices and edges are distinct. Then $R^\varphi(G) = (V^\varphi, E^\varphi, \omega)$ is called Rough $\zeta$ – labeling graph if it satisfies the following conditions:

i) $R^\varphi(G) = \max(\omega(v_i^\varphi), \omega(v_j^\varphi)) > 0$, edge exists .

ii) If $\sigma^\varphi(uv) = \frac{\zeta}{2}$, $\zeta$ is even and $\sigma^\varphi(uv) = \frac{\zeta+1}{2}$, $\zeta$ is odd for all $u,v \in V$ where $\zeta = f(u) + f(v) + m(G)$ for all $u,v \in V$

**3.4 Theorem:** The Rough path graph admits even vertex $\zeta$ –graceful labeling for all $n \geq 2$

**Proof:** Let $\mathcal{P}_n$ be a path rough graph with $n$ vertices and $n-1$ edges which represents in figure 1.

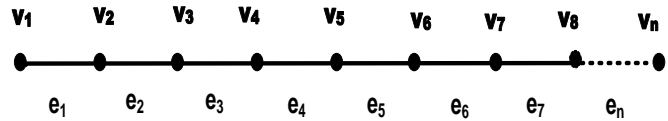

**Fig 1 :Rough path $\mathcal{P}_n$**

Defining the vertex labels with the function if

$f: V(G) \to \{2,4,6 \dots\}$ by $f(v_i) = 2i$ for $1 \leq i \leq n$. The edge labels are defined into two cases:

**Case (i) :** If $n$ is even then the mapping for the edge labeling is defined as $f: E(G) \to N$ with the function

$f(e_i) = \dfrac{n+5+j}{2}$ for $j = 4i - 3$ and $i = 1,2,3 \dots n-1$.

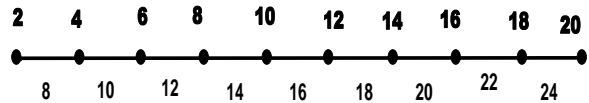

**Fig 2: Rough path $\mathcal{P}_{10}$**

**Case(ii):** If $n$ is odd then the edge labeling is defined as follows:

$f(e_i) = \dfrac{n+4+j}{2}$ for $j = 4i - 3$ and $i = 1,2, \dots n-1$.

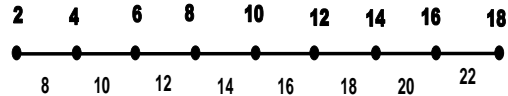

**Fig 3: Rough path $\mathcal{P}_9$**

**3.3 Theorem:** The rough cycle $C_n$ admits even vertex $\zeta$ −graceful labeling for all $n \geq 3$.

**Proof:** Let $v_1, v_2, \dots, v_n$ be the vertices of $C_n$ and $e_1, e_2, \dots, e_n$ be the edges of $C_n$. There are two cases:

**Case(i):** If $n$ is odd then the vertex labeling is defined as $f(v_i) = 2i$ for $1 \leq i \leq n$. The induced edge labels are as follows:

$f(e_i) = \dfrac{n+6+j}{2}$ where $\begin{cases} j = 4i - 3 & \text{for } i = 1,2, \dots n-1 \\ j = 2i - 3 & \text{for } i = n \end{cases}$

**Case(ii):** If $n$ is even then we define label of the vertices of $C_n$ as follows:

$f(v_i) = 2i$, $1 \leq i \leq n-1$,

$f(v_i) = 2i+2$, $i = n$.

There exists the induced edge labels are as follows:

$f(e_i) = \dfrac{n+5+j}{2}$ where $\begin{cases} j = 4i - 3, i = 1,2, \dots n-2 \\ j = 4i - 5, \quad i = n-1, n \end{cases}$

Hence the edges are distinct.

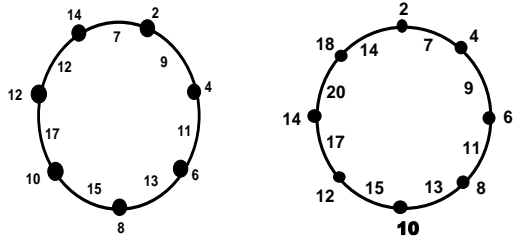

**Fig 4: Rough Cycle $C_7$ and $C_8$**

**3.4 Theorem**: The Rough star graph $S_{1,t}$ admits even vertex $\zeta$-graceful labeling for $t \geq 2$.

**Proof:** Let $G$ be a rough graph obtained by replacing each vertices of $S_{1,t}$ except the apex vertex $u_0$. It is the central vertex of the graph $G$. Let $u_i$ be the vertices of rough star graph for $1 \leq i \leq n$.

The Vertex labeling is $f(u_i) = 2i$ for $1 \leq i \leq n$.

The edge labeling is $f(u_0 u_i) = 2i$ and $f(u_0) = 0$.

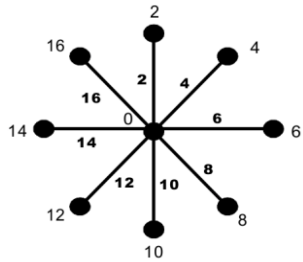

**Fig 5: Rough star for n=8**

**3.5 Theorem:** The rough comb $K_n$ admits even vertex $\zeta$ – graceful labeling for all $n \geq 3$.

**Proof:** Let $v_1, v_2, \ldots, v_n$ and $u_1, u_2, \ldots, u_n$ be the vertices of $K_n$. The general form is given in fig 6. The edge set is defined as follows:

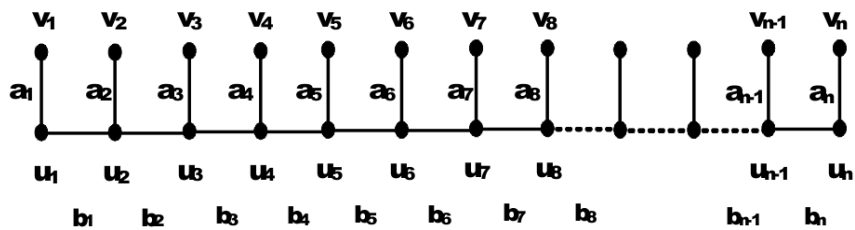

**Fig 6: Rough comb- General form**

**Case(i):** If $n$ is odd, define the vertex labeling is

$f(u_i) = 2i$   for $1 \leq i \leq n$, and

$f(v_i) = 2i + 2n + 2$ for $i = 1, 2, \ldots n$

Then the edge labeling is defined as

$f(u_i u_{i+1}) = n + 2i + 1$,

$f(u_i v_i) = 2n + 2i + 1$ for $i \in N$. This was shown in fig 7.

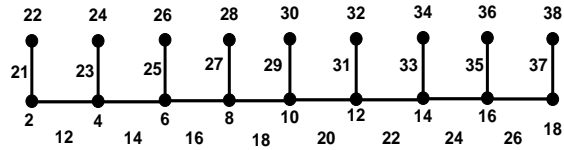

Fig 6:Rough comb- General form

Fig 7:Rough comb graph $K_9$

**Case(ii):** If $n$ is even then the vertex labeling is

$f((u_i)) = 2i$ for $1 \leq i \leq n$ and

$f(v_i) = 2i + 2n\ 1$ for $i \in N$.

The edge labeling is defined as follows:

$f(u_i u_{i+1}) = n + 2i + 1$

$f(u_i v_i) = 2n + 2i$ for $i \in N$.

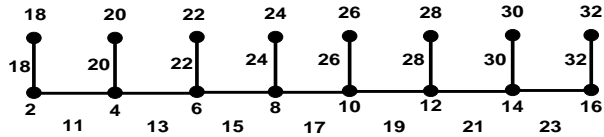

Fig 8:Rough comb graph for $K_8$

**3.6 Theorem:** The Rough Ladder graph $L_n$ admits even vertex $\zeta$-graceful labeling.

**Proof:** Let $V(L_n) = \{u_i, v_i / 1 \leq i \leq n\}$ be vertex set and

$E(L_n) = \{u_i u_{i+1}, v_i v_{i+1} / 1 \leq i \leq n-1\} \cup \{u_i v_i / 1 \leq i \leq n\}$ be the edge set of ladder $L_n$ then it has $2n$ vertices and $3n - 2$ edges as represented in fig 9.

Case(i): If $n$ is odd, then the vertex labeling is

$f(u_i) = 2i$ for $1 \leq i \leq n$,

$f(v_i) = 2i + 2n + 2$.

The edge labeling is defined as

$f(u_i u_{i+1}) = n + 2i + \left(\dfrac{n+1}{2}\right)$,

$f(u_i v_i) = 2n + 2i$

Case(ii): If $n$ is even then the vertex labeling is

$f(u_i) = 2i$ for $1 \leq i \leq n$, and

$f(v_i) = 2i + 2n\ 1$ for $i \in N$.

The edge labeling is

$f(u_i u_{i+1}) = n + 2i + 1$,

$f(u_i v_i) = 2n + 2i$ for $i \in N$.

Both the cases are explained in fig 10 and 11.

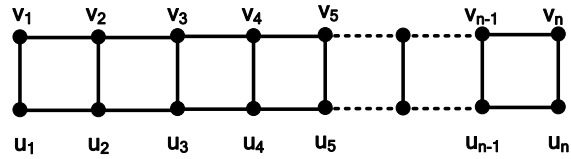

Fig 9: Rough ladder graph general form

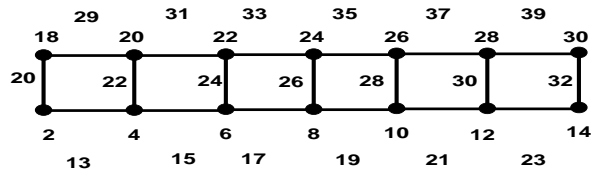

Fig 10: Rough ladder graph $L_7$

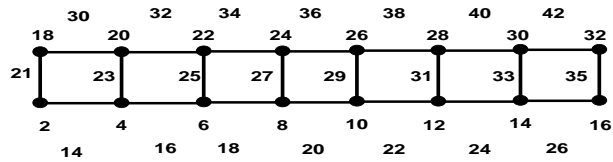

Fig 11: Rough ladder graph $L_8$

**3.7 Theorem:** The rough graph $P_n * K_{1,t}$ is $\zeta$-labeling.

**Proof:** Let G be a rough graph obtained by merging path graph and comb graph $P_n * K_{1,t}$ for t=2.

Let $V(P_n * K_{1,t}) = \{u_i, 1 \le i \le n\} \cup \{v_i, a_i, b_i / 1 \le i \le n\}$ be vertex set and

$E(P_n * K_{1,t}) = \{u_i, u_{i+1}/1 \le i \le n\} \cup \{u_i v_i, v_i a_i, v_i b_i, 1 \le i \le n\}$ be the edge set of rough graph. It was represented in the fig 12.

**Case (i):** If n is odd, then the vertex labeling is

$f(u_i) = 4i - 2,$

$f(v_i) = 4i,$

$f(a_i) = v_n - 2 + 4i,$

$f(b_i) = v_n + 4i.$

The edge labeling is defined as follows:

$f(u_i u_{i+1}) = 2n + 4i,$

$f(u_i, v_i) = 2n + 4i - 1,$

$f(v_i a_i) = f(a_i) + 1,$

$f(v_i b_i) = f(b_i).$ It was explained in fig 13.

**Case(ii):** If $n$ is even, then the vertex labeling is

$f(u_i) = 4i - 2,$

$f(v_i) = 4i,$

$f(a_i) = v_n + 2 + 4i,$

$f(b_i) = v_n + 4 + 4i.$

The edge labeling is defined as following:

$f(u_i u_{i+1}) = 2n + 4i,$

$f(u_i v_i) = 2n + 4i - 1,$

$f(v_i a_i) = f(a_i) - 1$

$f(v_i b_i) = f(b_i) - 2$

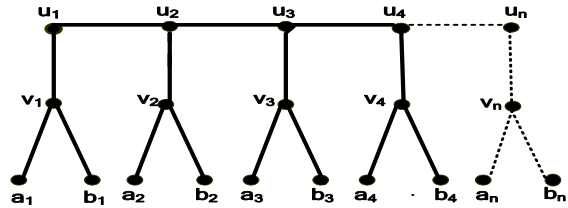

**Figure 12**

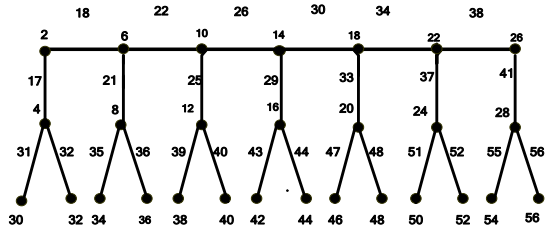

**Figure 13**

## Conclusion:

Rough graph has a wide application in computer networking. Labeling plays a vital role in graph theory. In this paper, we introduced a new type of labeling called even vertex $\zeta$-graceful labeling in rough graphs..Further we investigate this labeling for odd vertex for some special type of families of rough graphs in future work.